\documentclass[10pt,twocolumn,letterpaper]{article}

\usepackage{3dv}
\usepackage{times}
\usepackage{epsfig}
\usepackage{graphicx}
\usepackage{amsmath}
\usepackage{amssymb}

\usepackage{enumitem}
\usepackage{multirow}
\usepackage{multicol}
\usepackage{hhline}
\usepackage{pifont}
\newcommand{\xmark}{\ding{55}}%
\usepackage{booktabs}

\usepackage[pagebackref=true,breaklinks=true,colorlinks,bookmarks=false]{hyperref}

\threedvfinalcopy 

\def\threedvPaperID{****} 

\ifthreedvfinal\fi
\begin{document}

\title{SVNet: Where SO(3) Equivariance Meets Binarization on Point Cloud Representation}

\author{Zhuo Su$^{1,}$\thanks{The work was done when visiting AMLab, University of Amsterdam. $\dagger$ Correspondence to: Li Liu - \href{http://lilyliliu.com}{http://lilyliliu.com}.}\;\;\;\;\;
Max Welling$^2$\;\;\;\;\;
Matti Pietik{\"a}inen$^1$\;\;\;\;\;
Li Liu$^{3,1,\dagger}$\\
$^1$Center for Machine Vision and Signal Analysis, University of Oulu, Finland\\
$^2$AMLab, University of Amsterdam, Netherlands\\
$^3$National University of Defense Technology, China\\
{\tt\small zhuo.su@oulu.fi, m.welling@uva.nl, \{matti.pietikainen, li.liu\}@oulu.fi}
}

\maketitle
\thispagestyle{empty}

\begin{abstract}
    Efficiency and robustness are increasingly needed for applications on 3D point clouds, with the ubiquitous use of edge devices in scenarios like autonomous driving and robotics, which often demand real-time and reliable responses. The paper tackles the challenge by designing a general framework to construct 3D learning architectures with SO(3) equivariance and network binarization. However, a naive combination of equivariant networks and binarization either causes sub-optimal computational efficiency or geometric ambiguity. We propose to locate both scalar and vector features in our networks to avoid both cases. Precisely, the presence of scalar features makes the major part of the network binarizable, while vector features serve to retain rich structural information and ensure SO(3) equivariance. The proposed approach can be applied to general backbones like PointNet and DGCNN. Meanwhile, experiments on ModelNet40, ShapeNet, and the real-world dataset ScanObjectNN, demonstrated that the method achieves a great trade-off between efficiency, rotation robustness, and accuracy. The codes are available at  \href{https://github.com/zhuoinoulu/svnet}{https://github.com/zhuoinoulu/svnet}.
\end{abstract}

\section{Introduction}

3D point cloud processing has become a popular topic in recent years, with its broad applications like autonomous driving, augmented reality, and robotics. Deep neural networks are the first choices for building high accuracy architectures to recognize point cloud data, though often with huge computational cost and memory storage. Nowadays, applications on edge devices need more running efficiency and model compactness, meanwhile, rotation robustness, to deal with unseen environments with arbitrary poses in 3D data. Numerous efforts have been taken in tackling the challenge in either model efficiency by leveraging network binarization~\cite{qin2020bipointnet}, which has the charming advantages of up to 32$\times$ and $64\times$ reduction in memory storage and inference speed, respectively~\cite{rastegari2016xnor}, or rotation robustness by manipulating geometric features~\cite{zhang2019riconv,esteves2018sphericalcnns,zhang2020gc-conv}. We believe it can be solved within a single framework.

\begin{figure}[t!]
\centering
    \centering
    \includegraphics[width=0.96\linewidth]{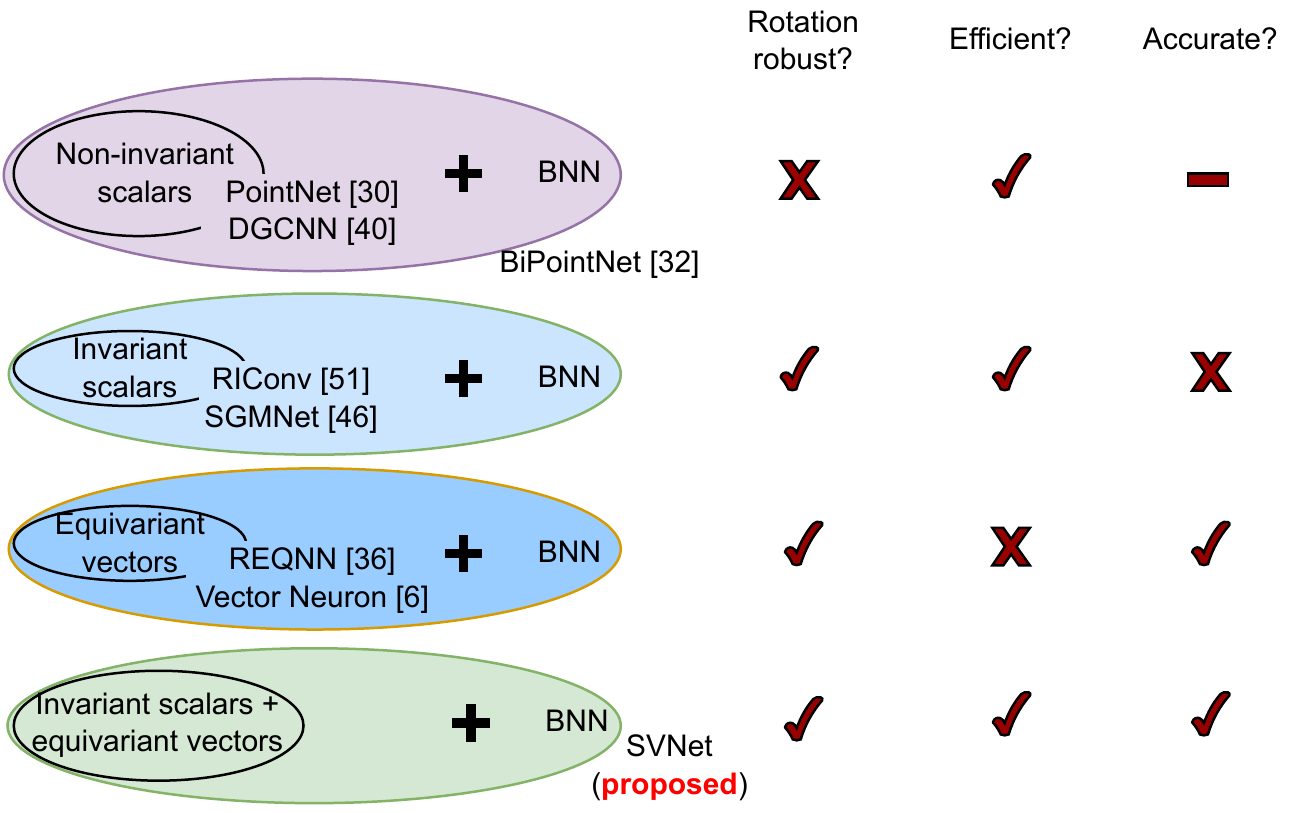}
    \caption{Binarization on different types of architectures.}
    \label{fig:combination}
\end{figure}

We start by discussing a naive combination with SO(3) invariant networks and network binarization by taking advantage of the latest approaches on both sides (Fig.~\ref{fig:combination}, Tab.~\ref{tab:snetvnet}). Generally, rotation-invariant networks utilize pose-preserving features during inference, which can be the rotation invariant ones~\cite{zhang2019riconv,xu2021sgmnet,li2021riframework,chen2019clusternet} in form of scalar geometric attributes like vector norms and angles, or the rotation equivariant ones~\cite{deng2021vn,shen20203reqnn} with vectors like coordinates or directions. On one hand, directly binarizing invariant scalar features and model weights still preserves rotation invariance and gives great speedup via replacing floating-point multiplications to the cheap XNOR-Count operations (or binary operations), while this combination may cause inevitable loss of geometric information~\cite{li2021riframework,xu2021sgmnet}, leading to unsatisfactory prediction accuracy. On the other hand, as binarizing equivariant vectors destroys rotation equivariance, merely binarizing weights and leaving those vectors unchanged gives another viable but sub-optimal option as the expensive floating-point multiplications are converted to additions, instead of the cheaper binary operations.


\begin{figure*}[t!]
\centering
    \centering
    \includegraphics[width=0.96\linewidth]{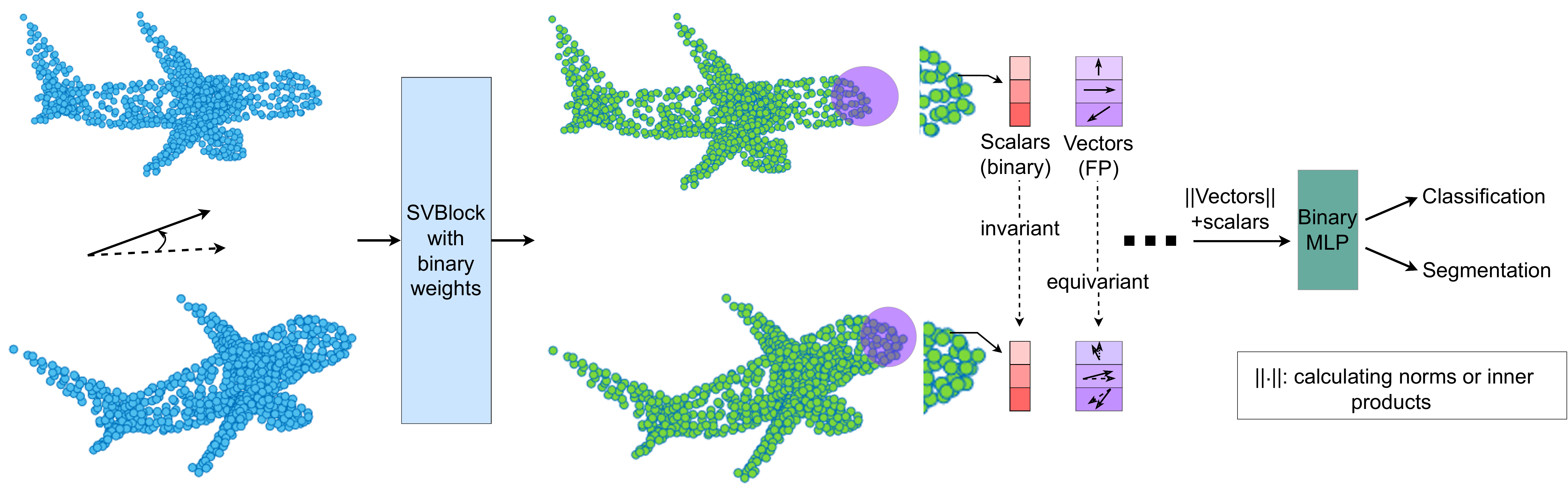}
    \caption{The overall structure of SVNet towards rigorous rotation invariance and model efficiency.}
    \label{fig:overview}
\end{figure*}

Motivated by the recent works on molecular representation learning~\cite{JingESTD21gvp,jing2021gvp2,schutt2021painn,satorras2021egnn}, where both invariant scalar features and equivariant vector features are updated and the two interact with each other during inference (in the rest of the paper, we use ``scalar'' and ``vector'' to represent ``invariant scalar feature'' and ``equivariant vector feature'' respectively for simplicity), we propose to build the 3D point cloud learning architectures in a similar way, namely, with dual use of scalars and vectors in feature updating. Despite the similar form, our motivation is different. For molecular processing, scalar and vector features naturally exist in data like bond types, electronegativity of the atoms (scalars), and atom orientations, edge directions (vectors), making it more intuitive to update both types of features. While it is not the case in 3D point clouds where usually only the coordinates (vectors) of points are given\footnote{invariant scalar features need to be manually created via geometric attributes like norms and inner products of vectors.}. Consequently there is a lack of investigation where both are used. We adopt scalars and vectors from the perspective of enhancing model efficiency when it comes to network binarization and rotation robustness: keeping vector features unchanged and binarizing scalars and model weights (Fig.~\ref{fig:overview}). The setting of dual use makes a difference compared with the conventional sole use of scalars or vectors, to particularly construct binary rotation invariant networks for point clouds. Firstly, scalars make a major part of the network fully ``binarizable'', which is key to gain high efficiency, and conveniently introduce nonlinearity for the model during inference, which provides another benefit as it is nontrivial to introduce nonlinearity for a pure vector-based equivariant model~\cite{deng2021vn,poulenard2021functional}. A certain amount of scalars on their own also enlarge the network capacity. Vector features, in contrast, makes an integral part to preserve the structural information without geometric ambiguities and ensure rotation equivariance. Although introducing a few addition operations from leaving vectors full-precision and binarizing weights, the vast majority of computations in the models are still made up of the cheapest binary operations due to the binarization of scalars. Finally, the invariance of the network is achieved by converting the equivariant vector features in the last layer to invariant ones. 

Based on the scalar-vector architectures, we can naively adopt existing binarization algorithms to achieve both efficiency and rotation invariance. The derived networks are dubbed as SVNet. As a ``by-product'', we also found the full-precision versions of SVNets provided state-of-the-art performances among prior ones which handle complicated geometric attributes~\cite{chen2019clusternet,zhang2019riconv,li2021riframework} or adopt pose disambiguation algorithms~\cite{yu2020deeppositional,kim2020ri-gcn,zhao2019lgr-net} for rotation invariance. It again indicates the potential of dual use of scalar and vectors for rotation-robust point cloud representation.

We summarize the contributions as follows:

\begin{enumerate}
    \item We proposed the dual use of invariant scalar and equivariant vector features for building efficient binary and rotation robust networks for 3D point clouds.
    \item We proposed a novel feature updating block for the case of dual use of scalars and vectors. The block is applicable to general backbones such as DGCNN~\cite{wang2019dgcnn} and PointNet~\cite{qi2017pointnet}, to enable high model efficiency and SO(3) equivariance.
    \item We conducted extensive experiments on widely used datasets, proving that our method can achieve comparable results with the complexity significantly reduced in terms of both memory storage and computational cost. The full-precision versions of the derived architectures also obtained state-of-the-art performance compared with the latest methods.
\end{enumerate}

\section{Related Work}



\vspace{0.3em}
\noindent \textbf{SO(3) symmetry on point clouds.} \quad
Both rotation invariance and equivariance can achieve SO(3) symmetry for deep neural networks on point clouds. On one hand, works on the former can leverage invariant geometric attributes like vector norms, distances, and relative angles to extract features~\cite{chen2019clusternet,zhang2019riconv,li2021riframework}. This kind of method often suffers from inevitable information loss caused by converting the vectors to scalars, where most of the positional information collapses. Another popular solution on invariance is to identify the canonical directions of the shape, via PCA~\cite{yu2020deeppositional,kim2020ri-gcn,li2021closer,zhang2020gc-conv} or SVD~\cite{zhao2019lgr-net}. Precisely predicting canonical directions can be well-tackled~\cite{atzmon2022frame}, this paradigm may suffer from the presence of potential geometric ambiguities which hinder their performances~\cite{li2021closer}. Clarification of possible ambiguities can also lead to linear growth in network complexity~\cite{li2021closer}. 
On the other hand, equivariant neural networks have been developed with the theory of group convolution~\cite{cohen2016group,cohen2018spherical} and steerable filters~\cite{cohen2016steerable,weiler20183dsteerablecnns}. An arbitrary 3D input can be projected to a unit sphere, or regular 3D voxels, and processed with group convolution either spatially or spectrally. The spatial-based methods~\cite{worrall2018cubenet,chen2021equivariantcvpr21} cannot easily guarantee equivariance on continuous groups and are often aided by extra data augmentation. In the spectral domain, the calculation of convolution involves the complicated Fourier transformation, and introducing nonlinearity becomes nontrivial~\cite{cohen2018spherical,esteves2018sphericalcnns}. We can also take advantage of the equivariance-preserving steerable filters which can be pre-defined like spherical harmonics~\cite{thomas2018tfn,poulenard2021functional} or learned~\cite{weiler20183dsteerablecnns,weiler2018learningsteerable}. While the filter kernels are not binarizable to preserve equivariance. A more practical and efficient way is vector mapping, where the three ``coordinates'' of vectors share the same weights/coefficients, as done in~\cite{deng2021vn,shen20203reqnn}. Vector mapping also enjoys the advantage that the weights are equivariantly binarizable.

\vspace{0.3em}
\noindent \textbf{Binary neural networks (BNNs).} \quad
BNNs were pioneered by Hubara \emph{et al.}~\cite{hubara2016bnn} for 2D image recognition using CNNs, where both activations and weights are binarized, thus achieving substantial network compression (up to 32$\times$ reduction in memory storage) and acceleration (up to 64$\times$ faster in inference speed~\cite{rastegari2016xnor}). The charming advantages of BNNs also introduced considerable attention for later researchers~\cite{rastegari2016xnor, liu2020reactnet, zhao2020bnnreview}. The most related work of ours in the BNN literature is BiPointNet~\cite{qin2020bipointnet}, which was possibly the first binarization approach to deep learning on point clouds. BiPointNet conducted binarization on the rotation sensitive structures like PointNet, without analysis or solutions on SO(3) equivariance. In our method, once the network is made binarizable with SO(3) equivariance, it is compatible with most of the BNN methods in the literature.


\vspace{0.3em}
\noindent \textbf{Others.} \quad
Our work also relates to the methods series on point clouds~\cite{qi2017pointnet,qi2017pointnet++,wang2019dgcnn,zhang2019shellnet,li2018pointcnn,wu2019pointconv}. Please refer to~\cite{guo2020pointcloudreview} for a more comprehensive review. At the same time, we are inspired by approaches on molecules like \emph{geometric vector perceptrons} on molecular structures where both scalar and vector features were utilized~\cite{schutt2021painn,JingESTD21gvp,jing2021gvp2}.

\section{Method}

\subsection{Problem statement and overall framework}
Without loss of generality, we introduce two example tasks on point clouds. Suppose $\mathcal{O}=\{o_1, o_2, ..., o_n\}$ is a given unordered point set, with each point assigned with a 3D coordinate: $o_i\in \mathbb{R}^3$. For the classification task, a network functions as a map: $\mathcal{O}\to \mathbb{Z}$, where $\mathbb{Z}$ is an integer representing the category of $\mathcal{O}$. In a part segmentation task, the network then turns to a dense prediction function: $(\mathcal{O},C)\to \mathbb{Z}^n$, where $C$ is the category of $\mathcal{O}$, by assigning each point $o_i$ to an integer that indicates which part of the original shape the point belongs to. The purpose of this paper is to make the network own the following properties:

\vspace{0.3em}
\noindent \textbf{SO(3) Equivariant:} \quad
Any basic layer $f^l$ in our network: $f^l(\mathcal{X}^l)=\mathcal{X}^{l+1}$, where $l$ is the layer index, $\mathcal{X}^l$ and $\mathcal{X}^{l+1}$ represent the input and output of this layer respectively, should meet:
\begin{equation}
    R^{l+1}_g\circ\mathcal{X}^{l+1} = f^l(R^l_g\circ\mathcal{X}^l)\;\;\;\; \forall g\in G,
    \label{eq:1}
\end{equation}
where $R^{l+1}$ and $R^l$ are the group representations of $G$ (in our paper, $R^{l+1}=R^l$ is a $3\times 3$ rotation matrix), and $R_g\circ\mathcal{X}$ a rotation $g$ on $\mathcal{X}$ as $G$ is the continuous SO(3) group. The above equation means that rotating the input first then passing it through the function $f^l$ gives the same result if we first pass it through $f^l$ and then rotate the output. 

Specifically, in the proposed SVNet, $\mathcal{X}$ is formed as $(\mathcal{S}, \mathcal{V})$ with $\mathcal{S}$ and $\mathcal{V}$ representing scalar and vector features respectively. Rotation on $\mathcal{X}$ is defined as:
\begin{equation}
    R_g\circ \mathcal{X} = R_g\circ (\mathcal{S}, \mathcal{V}) = (\mathcal{S}, R_g\mathcal{V}).
\end{equation}

A stacking of equivariant convolutional layers will result in an equivariant network (supposing the index of the last equivariant layer is $L$), as 
\begin{align}
    &f^L(...f^2(f^1(R^1_g\circ\mathcal{X}^1))) = f^L(...f^2(R^2_g\circ f^1(\mathcal{X}^1)))\nonumber\\
    &=f^L(...R^3_g\circ f^2(f^1(\mathcal{X}^1))) = ... \nonumber\\
    &= R^{L+1}_g\circ f^L(...f^2(f^1(\mathcal{X}^1))).
\end{align}

In SVNet, $\mathcal{V}^1$ can be created by using the original coordinates and relational positions between points. While $\mathcal{S}^1$ are generated from $\mathcal{O}$ via invariant geometric attributes (\emph{e.g.}, norms). The whole network can be made SO(3) invariant by converting the output $\mathcal{V}^L$ to invariant ones with the same strategy.

\vspace{0.3em}
\noindent \textbf{Efficient:} \quad
We take advantage of the BNNs to compose SVNet, where weights and scalar features are binarized, such that most of the multiplication operations can be circumvented by instead using much more efficient addition and binary operations. 

\begin{figure*}[t!]
\centering
    \centering
    \includegraphics[width=0.99\linewidth]{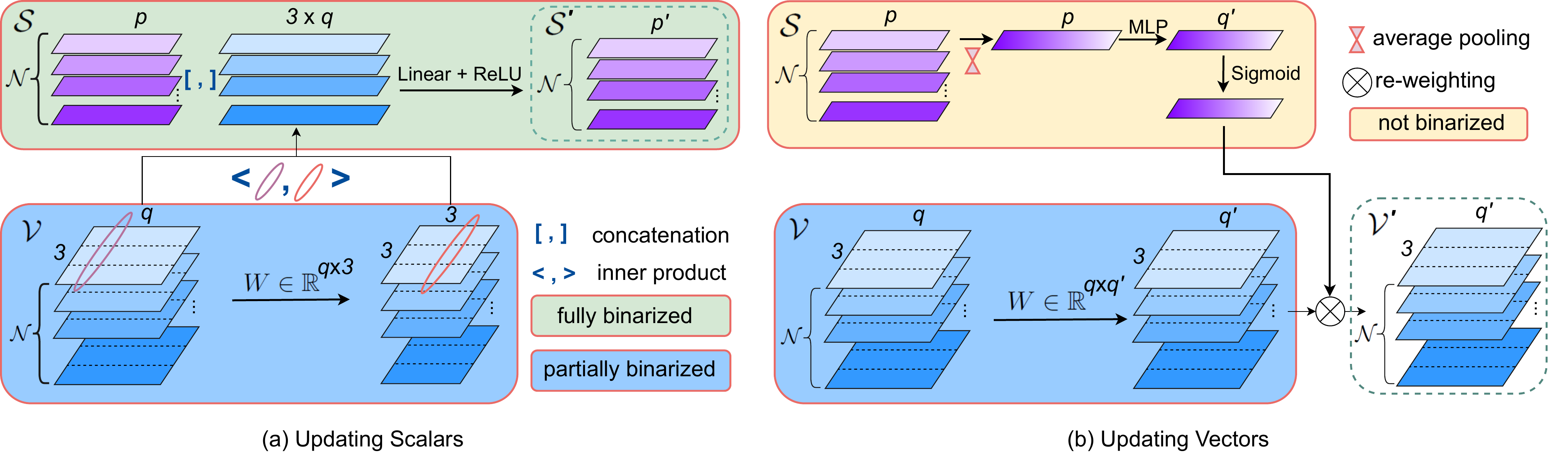}
    \caption{The SVBlock architecture. ``Fully binarized'' means both scalars and weights are binarized, ``partially binarized'' means keeping the features full-precision and binarizing the weights. The top right block is not binarized since it only introduces negligible computation due to average pooling.}
    \label{fig:svblock}
\end{figure*}

\subsection{Design of SVNet}
\label{sec:svnet}

Given an input $\mathcal{X}=(\mathcal{S}, \mathcal{V})$ with $\mathcal{S}\in\mathbb{R}^{p\times\mathcal{N}}$ and $\mathcal{V}\in\mathbb{R}^{3\times q\times\mathcal{N}}$, where $\mathcal{N}$ is the number of nodes or edges and $p$ ($q$) is the feature dimension (or the number of channels) for scalars (vectors), the core of SVNet is to design a ``binarizable'' convolutional layer where Eq.~\ref{eq:1} holds. 

\vspace{0.3em}
\noindent \textbf{Vector mapping} \quad
Generally, the invariant $\mathcal{S}$ can be freely manipulated with existing linear and nonlinear functions (including binarization). To equivariantly update vectors features $\mathcal{V}$, we can build linear mapping functions along the feature dimension $q$ and share the mapping along coordinates~\cite{shen20203reqnn,deng2021vn}.

As points are orderless, each $v\in\mathbb{R}^{3\times q}$ in $\mathcal{V}$ shares the same mapping functions for permutation equivariance.
Supposing the weight matrix is $W\in\mathbb{R}^{q\times q'}$, the linear mapping on $v$ is defined as:
\begin{equation}
    v' = f_v(v;W) = vW,
\end{equation}
transforming $v$ to $v'\in\mathbb{R}^{3\times q'}$ (and sequentially, $\mathcal{V}$ to $\mathcal{V}'\in\mathbb{R}^{3\times q'\times \mathcal{N}}$). It can be easily proved that $f_v$ is SO(3) equivariant: $R_gf_v(v;W) = R_g(vW) = (R_gv)W = f_v(R_gv;W)$. At the same time, binarization on $W$ does not affect this equivariance.

\vspace{0.3em}
\noindent \textbf{SVBlock: the basic building block} \quad
The design should also satisfy the following conditions. 
Firstly, proper nonlinear functions should be incorporated during the updating of $\mathcal{S}$ and $\mathcal{V}$ to make the network more discriminative. Secondly, $\mathcal{S}$ and $\mathcal{V}$ should interact with each other for better information fusion to strengthen network representation. The proposed SVBlock architecture is illustrated in Fig.~\ref{fig:svblock}, which we elaborate in detail as follows.

When updating scalar features, similar to~\cite{deng2021vn}, we first generate an equivariant coordinate system for each $v\in\mathcal{V}$ using vector mapping, and project $v$ to invariant features:
\begin{align}
    v_c = vW_c;\label{eq:5}\\
    v_{in} = v_c^Tv,\label{eq:6}
\end{align}
where $W_c\in\mathbb{R}^{q\times 3}$. $v_{in}$ is variant since $v_{in}=W_c^T(v^Tv)$, which can also be interpreted as a linear transformation of the covariance matrix of $v$. 
The updating of $\mathcal{S}$ can be implemented by concatenation with the obtained $\mathcal{V}_{in}$, followed by a multi-layer perceptron (MLP) module and nonlinear functions. The process is shown in Fig.~\ref{fig:svblock} (a).

When updating vector features, a problem we have to consider is that vector features $\mathcal{V}$ cannot be directly processed with nonlinear functions like ReLU or Sigmoid function. Although Vector Neurons~\cite{deng2021vn} proposed equivariant direction learning and projection, we find this can be addressed more efficiently in our situation, by generating nonlinear re-weighting factors from scalar features. The factors are used to multiply with vector features. Hence this also introduces interactions between scalars and vectors. 

Specifically, we again use vector mapping to linearly update $\mathcal{V}$, where the dimension of $\mathcal{V}$ changes from $q$ to the target dimension $q'$. Next, $\mathcal{S}$ is compressed along $\mathcal{N}$ using average pooling, resulting in $\mathcal{S}_{com}\in\mathbb{R}^{p}$, which is then used to generate the re-weighting factors for the linearly updated vectors by MLP and Sigmoid, as shown in Fig.~\ref{fig:svblock} (b).

The proposed SVBlock differs from \emph{geometric vector perceptrons}~\cite{JingESTD21gvp,jing2021gvp2} with the presence of data-dependent coordinate systems (Eq.~\ref{eq:5}) for converting $\mathcal{V}$ to invariant features, and the way to generate re-weighting factors which is made lightweight with average pooling.

\begin{table*}[t!]
\begin{center}
\setlength{\tabcolsep}{0.01\linewidth}
\resizebox*{0.8\linewidth}{!}{
\begin{tabular}{l|c|c|c|c|c}
\toprule
& Eq.~\ref{eq:5} & Eq.~\ref{eq:6} & Scalar updating & Re-weighting factors & Vector updating \\
\hline
Vanilla block & - & - & $\mathcal{N}C_1C_2$ (MACs) & - & -\\
\hline
SVBlock & $\frac{3}{2}\mathcal{N}C_1$ (ADDs) & $\frac{3}{2}\mathcal{N}C_1$ (MACs) & $\frac{1}{2}\mathcal{N}C_1C_2$ (BOPs) & $\frac{1}{12}C_1C_2$ (MACs) & $\frac{1}{12}\mathcal{N}C_1C_2$ (ADDs)\\
\midrule
\multicolumn{6}{c}{Total ($C_1=C_2=256, \mathcal{N}=1024$)}\\
\hline
Vanilla block & \multicolumn{5}{l}{67.1M MACs}\\
SVBlock (FP) & \multicolumn{5}{l}{39.9M MACs}\\
SVBlcok & \multicolumn{5}{l}{0.4M MACs + 6.0M ADDs + 33.6M BOPs}\\
\bottomrule
\end{tabular}
}
\end{center}
\caption{Computational cost of feature updating in a conventional network block (vanilla block) and SVBlock.}
\label{tab:block}
\end{table*}

\vspace{0.3em}
\noindent \textbf{Instantiating SVNet} \quad
SVNet is not a detailed network architecture but can be instantiated by plugging SVBlock into general backbones, making our method widely applicable. The pipeline of build SVNet is as follows.
\begin{enumerate}
    \item Extract $(\mathcal{S},\mathcal{V})$ for the first SVBlock: Given an unordered point set $\mathcal{O}=\{o_1, o_2, ..., o_n\}$ with $o_i\in\mathbb{R}^3$, we firstly find the $k$ nearest neighboring points $\{o_{ij}|j=1,2,...,k\}$ for each point $o_i$. The vector features of each $(o_i, o_{ij})$ pair is $v_{ij}=[o_i;o_{ij}-o_i]\in\mathbb{R}^{3\times 2}$. Based on that, we calculate the scalar features for the $(o_i, o_{ij})$ pair by putting $v_{ij}$ in Eq.~\ref{eq:5} and~\ref{eq:6}, getting $s_{ij}\in\mathbb{R}^6$ (by flattening $\mathbb{R}^{3\times 2}$ to $\mathbb{R}^6$). The vector features of $o_i$ is $v_i\in\mathbb{R}^{3\times 2\times k}$, and of $\mathcal{O}$ is $\mathcal{V}\in\mathbb{R}^{3\times 2\times (k\times n)}$. Similarly, the scalar features of $\mathcal{O}$ is $\mathcal{S}\in\mathbb{R}^{6\times (k\times n)}$. We can regard $(k\times n)$ as a single dimension.
    
    \item Aggregation: Aggregation of $\mathcal{S}$ or $\mathcal{V}$ can be done by pooling along the $k$ dimension, resulting in $\mathcal{S}\in\mathbb{R}^{p\times n}$ and $\mathcal{V}\in\mathbb{R}^{3\times q\times n}$. In DGCNN, we can use the same strategy in 1 to extend the ``$k$'' dimension back (please also refer to the original paper~\cite{wang2019dgcnn}).
    
    \item After the last SVBlock: $\mathcal{V}$ is converted to invariant features using Eq.~\ref{eq:5} and~\ref{eq:6} again, which is then concatenated with $\mathcal{S}$. The network ends with a binary MLP with nonlinear functions according to the backbone architecture. In this step, we can safely binarize both features and weights as there are only scalar features.
\end{enumerate}

\vspace{0.3em}
\noindent \textbf{Binarization on SVNet} \quad
It should be noted that the main focus of the paper is to propose an effective SO(3) equivariant framework from the perspective of network binarization. In other words, we enable SVNet to be compatible with previous BNN techniques. For simplicity, we binarize SVNet based on the following equations and use STE~\cite{hinton2012ste} for gradient calculation (please see Appendix A for a more detailed description). With a more powerful binarization method applied, the network can be further enhanced:

\begin{align}
    Y &= \text{Sign}(X)\cdot\text{Sign}(W)\; &\text{if both } X\text{ and }W\text{ are binarized},\label{eq:7}\\
    Y &= X\cdot\text{Sign}(W)\;&\text{if only }W \text{ is binarized},\label{eq:8}
\end{align}
where $W$, $X$, and $Y$ represents the weights, input, and output respectively, $\text{Sign}(a) = +1$ if $a \geq 0$ otherwise -1, ``$\cdot$'' means matrix multiplication. 

In SVNet, Eq.~\ref{eq:7} corresponds to the linear transformation of scalar features while Eq.~\ref{eq:8} the vector mapping of vector features. The former can be implemented with BOPs (binary operations) and the latter with ADDs (additions), both are much more efficient than the original MACs (Multiply-Accumulates) in a full-precision counterpart.

%

\vspace{0.3em}
\noindent \textbf{Why is SVNet efficient?} \quad
We give an example on comparing the computational cost between a vanilla network block, where only scalar features are used) in conventional methods~\cite{qi2017pointnet,wang2019dgcnn}, and SVBlock, by analyzing the complexity of feature updating. For the former, the scalar features are updated with function: $\mathbb{R}^{C_1\times \mathcal{N}}\to\mathbb{R}^{C_2\times \mathcal{N}}$. To keep the same amount of features, SVBlock functions as: $(\mathbb{R}^{\frac{C_1}{2}\times\mathcal{N}}, \mathbb{R}^{3\times \frac{C_1}{6}\times\mathcal{N}})\to (\mathbb{R}^{\frac{C_2}{2}\times\mathcal{N}}, \mathbb{R}^{3\times \frac{C_2}{6}\times\mathcal{N}})$. As shown in Tab.~\ref{tab:block}, the existence of both scalar and vector features makes the full-precision SVBlock (FP) more compact than a vanilla counterpart (39.9M MACs \emph{vs.} 67.1M MACs). With binarization, most of the MACs is replaced with BOPs in SVBlock, which is the key to achieve better efficiency.

\section{Experiments}

In this section, we evaluate the proposed SVNet on the tasks of 3D object classification and part segmentation, based on the widely used ModelNet40~\cite{wu2015modelnet40}, ShapeNet~\cite{yi2016shapenet}, and ScanObjectNN~\cite{uy2019scanobjectnn} datasets. To show its rotation robustness and compare it with other methods, we follow the training/testing settings as in prior literature: $z$/$z$, $z$/SO3, and SO3/SO3, where $z$ and SO3 mean ``random rotation around $z$ axis'' and ``random rotation'' respectively. All the experiments were conducted with Pytorch library~\cite{paszke2019pytorch}. We adopt batch normalization~\cite{IoffeS15bn} following~\cite{deng2021vn,shen20203reqnn} and keep the first and last layer of SVNet full-precision to avoid severe information loss. Please also refer to Appendix B for a more challenging scenario where there was no rotation during training but was during testing.


\subsection{Experiments on ModelNet40}
\label{sec:modelnet40}

ModelNet40~\cite{wu2015modelnet40} has been extensively used for synthetic shape classification, including 12,311 CAD models with 40 man-made object categories (\emph{e.g.}, airplane, bathtub, \emph{etc.}). The dataset was split into 9,843 models for training and 2,468 for testing. To make a fair comparison with prior methods, we randomly extracted 1,024 3D points from each model for both training and testing. SVNet was instantiated with DGCNN~\cite{wang2019dgcnn} and PointNet~\cite{qi2017pointnet} backbones. Following~\cite{deng2021vn}, it was trained for 250 epochs for the DGCNN backbone with a cosine annealing learning rate~\cite{LoshchilovH17cosine}, and 200 epochs for the PointNet backbone with a multi-step annealing learning rate by decaying ($\times 0.7$) in every 20 epochs. For both cases, the learning rate was initialized as 0.001 and decayed towards 0, batch size was 32, and the Adam optimizer~\cite{kingma2014adam} was adopted. We also trained a full-precision version of our model, namely SVNet (FP) for each backbone, for a better comparison. 

\begin{table}[t!]
\begin{center}
\setlength{\tabcolsep}{0.005\linewidth}
\resizebox*{\linewidth}{!}{
\begin{tabular}{l|lcccc}
\toprule
& Method & Binarized & $z$/$z$ & $z$/SO(3) & SO(3)/SO(3)\\
\midrule
\multirow{6}{*}{\begin{tabular}{@{}c@{}}Rotation\\sensitive\end{tabular}} & PointNet$^*$~\cite{qi2017pointnet} & \xmark & 85.9 & 17.0 & 74.7 \\
& DGCNN~\cite{wang2019dgcnn} & \xmark & 90.3 & 33.8 & 88.6 \\
& PointNet++~\cite{qi2017pointnet++} & \xmark & 91.8 & 28.4 & 85.0 \\
& PointCNN~\cite{li2018pointcnn} & \xmark & 92.5 & 41.2 & 84.5 \\
& ShellNet~\cite{zhang2019shellnet} & \xmark & 93.1 & 19.9 & 87.8\\
& BiPointNet$^*$~\cite{qin2020bipointnet} & \checkmark & 39.9 & 13.7 & 16.6\\
\midrule
\multirow{8}{*}{\begin{tabular}{@{}c@{}}Rotation\\invariant\end{tabular}} & RIConv~\cite{zhang2019riconv} & \xmark & 86.5 & 86.4 & 86.4 \\
& ClusterNet~\cite{chen2019clusternet} & \xmark & 87.1 & 87.1 & 87.1 \\
& Yu \emph{et al.}~\cite{yu2020deeppositional} & \xmark & 89.2 & 89.2 & 89.2 \\
& RI-GCN~\cite{kim2020ri-gcn} & \xmark & 89.5 & 89.5 & 89.5 \\
& GC-Conv~\cite{zhang2020gc-conv} & \xmark & 89.0 & 89.1 & 89.2 \\
& Li \emph{et al.}~\cite{li2021riframework} & \xmark & 89.4 & 89.4 & 89.3 \\
& SGMNet~\cite{xu2021sgmnet} & \xmark & 90.0 & 90.0 & 90.0\\
& Li \emph{et al.}~\cite{li2021closer} & \xmark & 90.2 & 90.2 & 90.2 \\
\midrule
\multirow{4}{*}{\begin{tabular}{@{}c@{}}Rotation\\equivariant\end{tabular}} & Spherical CNNs~\cite{esteves2018sphericalcnns} & \xmark & 88.9 & 78.6 & 86.9 \\
& $\alpha^3$SCNN~\cite{liu2019a3scnn} & \xmark & - & - & 88.7\\
& Poulenard \emph{et al.}~\cite{poulenard2021functional} & \xmark & 90.5 & 88.2 & 89.3 \\
& VN-DGCNN~\cite{deng2021vn} & \xmark & 89.5 & 89.5 & 90.2 \\
\midrule
\multirow{4}{*}{\begin{tabular}{@{}c@{}c@{}}Rotation\\equivariant\\(\textbf{Proposed})\end{tabular}} & SVNet-PointNet (FP) & \xmark & 86.3 & 86.3 & 86.6\\
& SVNet-DGCNN (FP) & \xmark & 90.3 & 90.3 & 90.0\\
& SVNet-PointNet & \checkmark  & 76.3 & 76.3 & 75.8\\
& SVNet-DGCNN & \checkmark & 83.8 & 83.8 & 83.8\\
& SVNet-DGCNN$^\dagger$ & \checkmark & 86.8 & 86.8 & 86.8\\
\bottomrule
\end{tabular}
}
\end{center}
\caption{Comparison on ModelNet40. $^\dagger$ indicates a two-step training scheme. ``$*$'' indicates our implementations using the code provided by the authors. Numbers show overall accuracies (\%).}
\label{tab:modelnet40}
\end{table}

In Tab.~\ref{tab:modelnet40}, SVNet was compared with prior state-of-the-art methods including the rotation-sensitive, rotation-invariant, and rotation-equivariant ones. We had two main findings. Firstly, the full-precision version SVNet (FP) achieves 90.3\% accuracy, surpassing the latest Vector Neurons~\cite{deng2021vn} with 89.5\% accuracy via the sole use of vectors, and SGMNet~\cite{xu2021sgmnet} with 90.0\% accuracy via pure scalars, validating the effectiveness of the proposed scalar-vector configuration. Secondly, when binarized, the complexity of the network is significantly reduced in both memory and computational cost (Tab.~\ref{tab:complexity}). Nevertheless, SVNet still retains rigorous rotation invariance with competitive prediction performance. For instance, with our vanilla binarization, it achieves 83.8\% accuracy regardless of the training/testing settings, \emph{vs.} 39.9\%, 13.7\%, and 16.6\% by BiPointNet~\cite{qin2020bipointnet}. In addition, with a more powerful binarization technique\footnote{A two-step binarization, by firstly training the network with full-precision weights and activations, then training with binary weights and activations.}, SVNet achieves 86.8\% accuracy, being highly comparable with most of the prior works which adopted full-precision weights and activations.

\begin{table}[t!]
\begin{center}
\setlength{\tabcolsep}{0.01\linewidth}
\resizebox*{\linewidth}{!}{
\begin{tabular}{l|lccc}
\toprule
& Method & Binarized & $z$/SO(3) & SO(3)/SO(3)\\
\midrule
\multirow{5}{*}{\begin{tabular}{@{}c@{}}Rotation\\sensitive\end{tabular}} & PointNet~\cite{qi2017pointnet} & \xmark & 41.8 & 62.3 \\
& DGCNN~\cite{wang2019dgcnn} & \xmark & 49.3 & 78.6 \\
& PointNet++~\cite{qi2017pointnet++} & \xmark & 48.3 & 76.7 \\
& PointCNN~\cite{li2018pointcnn} & \xmark & 34.7 & 71.4 \\
& ShellNet~\cite{zhang2019shellnet} & \xmark & 47.2 & 77.1 \\
& BiPointNet~\cite{qin2020bipointnet} & \checkmark & 31.4 & 36.0\\
\midrule
\multirow{6}{*}{\begin{tabular}{@{}c@{}}Rotation\\invariant\end{tabular}} & RIConv~\cite{zhang2019riconv} & \xmark & 75.3 & 75.5 \\
& RI-GCN~\cite{kim2020ri-gcn} & \xmark & 77.2 & 77.3 \\
& Li \emph{et al.}~\cite{li2021riframework} & \xmark & 79.2 & 79.4 \\
& SGMNet~\cite{xu2021sgmnet} & \xmark & 79.3 & 79.3 \\
& Li \emph{et al.}~\cite{li2021closer} & \xmark & 81.7 & 81.7 \\
\midrule
\multirow{2}{*}{\begin{tabular}{@{}c@{}}Rotation\\equivariant\end{tabular}} & Poulenard \emph{et al.}~\cite{poulenard2021functional} & \xmark & 78.1 & 78.2 \\
& VN-DGCNN~\cite{deng2021vn} & \xmark & 81.4 & 81.4 \\
\midrule
\multirow{5}{*}{\begin{tabular}{@{}c@{}c@{}}Rotation\\equivariant\\(\textbf{Proposed})\end{tabular}} & SVNet-PointNet (FP) & \xmark & 78.2 & 78.6 \\
& SVNet-DGCNN (FP) & \xmark & \underline{81.4} & \underline{81.4} \\
& SVNet-PointNet & \checkmark  & 67.3 & 67.3 \\
& SVNet-DGCNN & \checkmark & 68.4 & 68.9 \\
& SVNet-DGCNN$^\dagger$ & \checkmark & 71.5 & 71.5\\
\bottomrule
\end{tabular}
}
\end{center}
\caption{Comparison on ShapeNet. Numbers show results on mean intersection of union (mIoU, \%) over all classes. Underline numbers in our submission version were both 80.9 as we incidentally used label smoothing, which were unnecessary.}
\label{tab:shapenet}
\end{table}

\subsection{Experiments on ShapeNet}
\label{sec:shapenet}

We used the ShapeNet part dataset~\cite{yi2016shapenet}  to evaluate our method on the part segmentation task. The dataset consists of 16,881 shapes with annotations of 50 parts in total, from 16 categories. It was split into 14,007 and 2,874 shapes for training and testing respectively. Different from ModelNet40, we utilized 2,048 points for each shape during training and testing. Again, we built SVNet via DGCNN and PointNet backbones. For both cases, SVNet was trained for 200 epochs with batch size of 32, using Adam optimizer~\cite{kingma2014adam} with an initial learning rate of 0.001. The learning rate was decayed to 0 with cosine annealing~\cite{LoshchilovH17cosine} for DGCNN and multi-step annealing with step size of 20 and decaying rate of 0.5 for PointNet. 

\begin{table*}[t!]
\begin{center}
\setlength{\tabcolsep}{0.01\linewidth}
\resizebox*{0.98\linewidth}{!}{
\begin{tabular}{l|l||c|ccc|c||c|ccc|c}
\toprule
\multicolumn{2}{c}{ } & \multicolumn{5}{c}{ModelNet40} & \multicolumn{5}{c}{ShapeNet}\\
\midrule
& Method & Params & MACs & ADDs & BOPs & $z$/SO(3) & Params & MACs & ADDs & BOPs & $z$/SO(3)\\
\hline
\multirow{4}{*}{DGCNN} & Original~\cite{wang2019dgcnn} & 57.7M & 2.4B & 0 & 0 & 33.8 & 46.7M & 4.4B & 0 & 0 & 49.3\\
& Vector Neurons~\cite{deng2021vn} & 92.8M & 3.2B & 0 & 0 & 89.5 & 41.8M & 6.6B & 0 & 0 & 81.4\\
& SVNet (FP) & 49.7M & 1.4B & 0 & 0 & 90.3 & 43.2M & 7.2B & 0 & 0 & 81.4\\
& \textbf{SVNet} & \textbf{3.4M} & \textbf{0.05B} & \textbf{0.2B} & \textbf{1.2B} & \textbf{83.8} & \textbf{4.0M} & \textbf{0.2B} & \textbf{1.0B} & \textbf{6.0B} & \textbf{68.4}\\
\midrule
\multirow{6}{*}{PointNet} & Original~\cite{qi2017pointnet}$^*$ & 111.1M & 0.4B & 0 & 0 & 17.0 & 267.0M & 5.8B & 0 & 0 & 41.8\\
& Vector Neurons$^*$~\cite{deng2021vn} & 63.1M & 2.0B & 0 & 0 & 85.6 & 162.6M & 20.5B & 0 & 0 & 79.8\\
& SVNet (FP) & 78.8M & 1.5B & 0 & 0 & 86.3 & 234.8M & 14.2B & 0 & 0 & 78.2\\
\hhline{~-----------}
& BiPointNet$^*$~\cite{qin2020bipointnet} & 4.2M & 0.01B & 0 & 0.4B & 13.7 & 9.0M & 0.1B & 0 & 5.7B & 31.4\\ 
& \textbf{SVNet} & \textbf{8.7M} & \textbf{0.03B} & \textbf{0.2B} & \textbf{1.2B} & \textbf{76.3} & \textbf{14.0M} & \textbf{0.2B} & \textbf{0.2B} & \textbf{13.8B} & \textbf{67.3}\\
& \textbf{SVNet-small} & \textbf{4.3M} & \textbf{0.02B} & \textbf{0.03B} & \textbf{0.2B} & \textbf{66.4} & \textbf{7.2M} & \textbf{0.1B} & \textbf{0.2B} & \textbf{5.2B} & \textbf{64.3}\\
\bottomrule
\end{tabular}
}
\end{center}
\caption{Complexity comparison on ModelNet40 and ShapeNet. The numbers show overall accuracies (\%) for ModelNet40 and mIoU (\%) for ShapeNet. Params (memory storage of the model) were recorded in bits. We also constructed SVNet-small to compare with BiPointNet with similar complexity. ``$*$'' indicates our implementations using the code provided by the authors.}
\label{tab:complexity}
\end{table*}

The experimental results and comparison with prior state-of-the-art methods are given in Tab.~\ref{tab:shapenet}. Similar to~\ref{sec:modelnet40}, we also give a comparison of network complexity in Tab.~\ref{tab:complexity}. We can get the consistent observations as in~\ref{sec:modelnet40}: the full-precision SVNet with DGCNN backbone achieves comparable results beating most of the prior rotation invariant/equivariant methods; equipped with SVBlock, the original backbones can be effectively improved in terms of rotation robustness; and finally, under binarization, SVNet shows a considerable reduction in memory and computation with high predictive results. Similarly, the performance can be further boosted with extra binarization technique as on ModelNet40.


\subsection{Experiments on ScanObjectNN}
\label{sec:scanobjectnn}

\begin{table}[t!]
\begin{center}
\setlength{\tabcolsep}{0.02\linewidth}
\resizebox*{\linewidth}{!}{
\begin{tabular}{l|lccc}
\toprule
& Method & Binarized & $z$/$z$ & $z$/SO(3)\\
\midrule
\multirow{4}{*}{Rotation-sensitive} & PointNet~\cite{qi2017pointnet} & \xmark & 68.2 & 17.1 \\
& DGCNN~\cite{wang2019dgcnn} & \xmark & 78.1 & 16.1 \\
& PointNet++~\cite{qi2017pointnet++} & \xmark & 77.9 & 15.8 \\
& PointCNN~\cite{li2018pointcnn} & \xmark & 78.5 & 14.9 \\
\midrule
\multirow{2}{*}{Rotation-robust} & RIConv~\cite{zhang2019riconv} & \xmark & 67.9 & 67.9 \\
& LGR-Net~\cite{zhao2019lgr-net} & \xmark & 72.7 & 72.7 \\
\hline
\multirow{3}{*}{\begin{tabular}{@{}c@{}}Rotation-robust\\(\textbf{Proposed})\end{tabular}} & \textbf{SVNet (FP)} & \xmark & \textbf{76.2} & \textbf{76.2} \\
& \textbf{SVNet} & \checkmark & \textbf{52.9} & \textbf{52.9} \\
& \textbf{SVNet}$^\dagger$ & \checkmark & \textbf{60.9} & \textbf{60.9} \\
\bottomrule
\end{tabular}
}
\end{center}
\caption{Comparison on ScanObjectNN with the setting \textbf{PB\_T50\_RS}. Numbers show overall accuracies (\%).}
\label{tab:scanobjectnn}
\end{table}

ScanObjectNN~\cite{uy2019scanobjectnn} is a real-world dataset having 2,902 object with 15 categories. Different from synthetic ones like ModelNet40~\cite{wu2015modelnet40}, it also contains background points, making classification much more challenging. To evaluate our method, we adopted the hardest setting in this dataset: \textbf{PB\_T50\_RS}, which consists of 14,298 objects revised from the base ones, with 11,416 and 2,882 objects for training and testing, respectively. We adopted the same training settings in~\ref{sec:modelnet40} using DGCNN backbone. The results are shown in Tab.~\ref{tab:scanobjectnn}. The network complexity of SVNet is almost the same as in Tab.~\ref{tab:complexity} with the DGCNN backbone, with the only difference in the last linear layer due to different numbers of categories.

\subsection{Ablation and visualization}
\label{sec:ablation}

If not specified, the following studies were conducted using the DGCNN backbone on the ModelNet40~\cite{wu2015modelnet40} dataset. The training settings followed Section~\ref{sec:modelnet40}.

\vspace{0.3em}
\noindent \textbf{How do we need S and V?} \quad
In our experiments, we equally divided scalar and vector features (\emph{i.e.}, $\mathcal{S}\in\mathbb{R}^{\frac{C}{2}\times \mathcal{N}}$ and $\mathcal{V}\in\mathbb{R}^{3\times\frac{C}{6}\times\mathcal{N}}$), so that both have the same amount of features. One may consider: \emph{What if we only use scalar features or vector features? What if we change this proportion?} We conducted such exploration as illustrated in Tab.~\ref{tab:snetvnet}, with different scalar-vector proportions. One design principle in this study is to make the complexity of other SVNet variants approximate or higher than the one adopted in our paper, and observe if the adopted one still achieves equal or better performance. We show how this study matches our original motivation.

\begin{table}[t!]
\begin{center}
\setlength{\tabcolsep}{0.01\linewidth}
\resizebox*{\linewidth}{!}{
\begin{tabular}{cll|cccc|c}
\toprule
S : V & Property & & Params & MACs & ADDs & BOPs & $z$/SO(3) \\
\hline
\multirow{2}{*}{$1:0$} & \multirow{2}{*}{invariant} & FP & 58.0M & 2.5B & 0 & 0 & 84.8\\
& & Binary & 2.5M & 0.1B & 0 & 2.4B & 71.7\\
\midrule
\multirow{2}{*}{$\frac{2}{3}:\frac{1}{9}$} & \multirow{2}{*}{equivariant} & FP & 52.3M & 1.7B & 0 & 0 & 90.3\\
& & Binary & 3.2M & 0.05B & 0.1B & 1.6B & 83.4 \\
\midrule
\multirow{2}{*}{$\star\;\frac{\mathbf{1}}{\mathbf{2}}:\frac{\mathbf{1}}{\mathbf{6}}$} & \multirow{2}{*}{equivariant} & FP & 49.7M & 1.4B & 0 & 0 & 90.3\\
& & Binary & 3.4M & 0.05B & 0.2B & 1.2B & 83.8\\
\midrule
\multirow{2}{*}{$0:\frac{1}{3}$} & \multirow{2}{*}{equivariant} & FP & 60.7M & 1.4B & 0 & 0 & 89.2\\
& & Binary & 11.0M & 0.08B & 1.3B & 0.001B & 83.4\\
\bottomrule
\end{tabular}
}
\end{center}
\caption{SVNet variants with different scalar-vector proportions on feature channels. ``$\star$'' indicates the finally adopted one in the paper. See Tab.~\ref{tab:complexity} for other notes.}
\label{tab:snetvnet}
\end{table}

Specifically, the proposed SVNet can be regarded as a general form of architectures in prior approaches, where only scalar or vector features are used. One case is the pure invariant version of SVNet (corresponding to $S:V=1:0$ in Tab.~\ref{tab:snetvnet}). This can be implemented by converting all the vectors (from the original coordinates and relational positions of points) to scalars using Eq.~\ref{eq:5} and~\ref{eq:6}. Though most of the computation can be implemented with BOPs, the invariant SVNet suffers from a considerable accuracy degradation. This simple generation of scalar features causes severe loss of information on geometric structures which leads to potential ambiguities. To solve this, prior methods take nontrivial effort on generating invariant geometric attributes from different perspectives like angles and distances~\cite{chen2019clusternet,zhang2019riconv}, or by considering global information~\cite{zhang2020gc-conv}.
While we found this problem can be trivially alleviated by introducing more vector features, as shown in the third and fourth rows of Tab.~\ref{tab:snetvnet}. 

The other extreme case of SVNet is when we completely discard scalar features in SVBlock (corresponding to the last row of the table). This reduces our method to ones like Vector Neurons~\cite{deng2021vn} or REQNN~\cite{shen20203reqnn}. The performance in this case is competitive. While compared with the ``standard'' SVNet we adopted in our experiments, it suffers from a sub-optimal level of efficiency as most of its computation turns to ADDs, rather than BOPs.

\begin{figure}[t!]
\centering
    \centering
    \includegraphics[width=\linewidth]{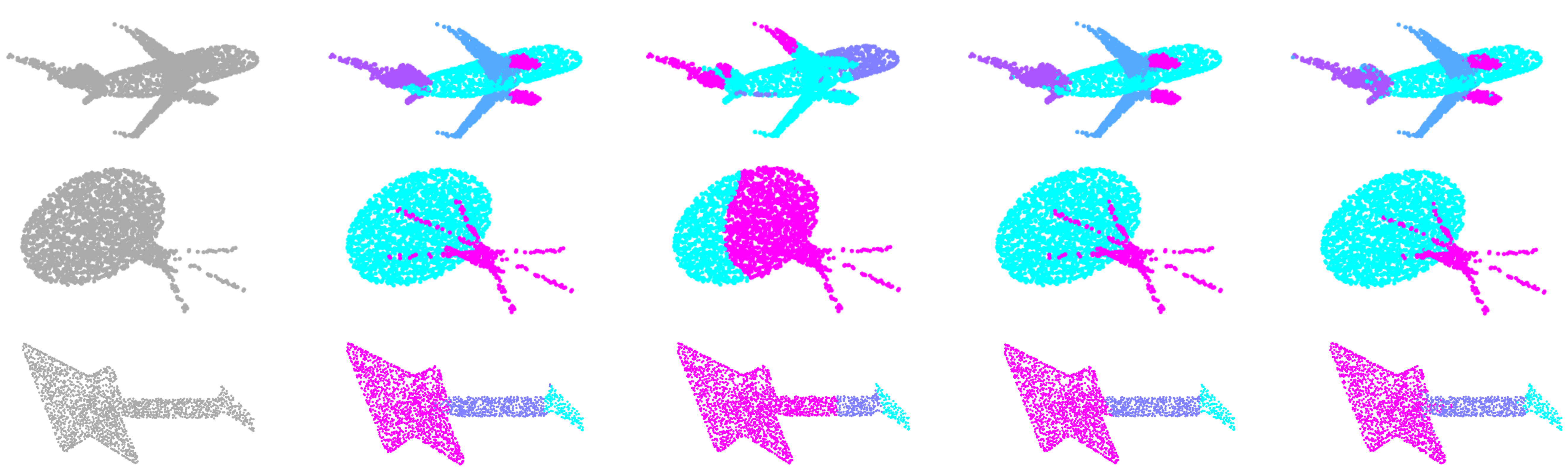}
    \caption{Visualization of part segmentation predictions on ShapeNet testing data using different structures. The input is randomly rotated before being fed to the models. Top to bottom: airplane, table, guitar. Left to right: input, ground truth, results from PointNet~\cite{qi2017pointnet}, results from SVNet-PointNet (FP), results from SVNet-PointNet}
    \label{fig:visualization}
\end{figure}

\begin{table}[t!]
\begin{center}
\setlength{\tabcolsep}{0.02\linewidth}
\resizebox*{0.95\linewidth}{!}{
\begin{tabular}{lccc}
\toprule
& \begin{tabular}{@{}c@{}}Scalar\\concatenation\end{tabular} & \begin{tabular}{@{}c@{}}Vector\\re-weighting\end{tabular} & \begin{tabular}{@{}c@{}}Overall accuracy\\($z$/SO(3), \%)\end{tabular}\\
\midrule
\multirow{4}{*}{Full-precision} & \xmark & \xmark & 88.8\\
& \xmark & \checkmark & 89.0\\
& \checkmark & \xmark & 90.8\\
& \checkmark & \checkmark & 90.3\\
\midrule
\multirow{4}{*}{Binary} & \xmark & \xmark & 79.5\\
& \xmark & \checkmark & 79.5\\
& \checkmark & \xmark & 81.5\\
& \checkmark & \checkmark & 83.8\\
\bottomrule
\end{tabular}
}
\end{center}
\caption{Influences of different components in SVBlock.}
\label{tab:interaction}
\end{table}

\vspace{0.3em}
\noindent \textbf{Feature interactions in SVBlock} \quad
Another consideration is the effectiveness of interactions between scalars and vectors during feature updating. In Tab.~\ref{tab:interaction}, we validated the two interaction modules in SVBlock: scalar concatenation and vector re-weighting, as introduced in~\ref{sec:svnet}. Both SVNet (FP) and SVNet were investigated since we wanted to see if there are any differences in full-precision and binary structures. The observations are generally consistent. For both structure types, interactions lead to better performances, especially for binary structures. The only exception is for the full-precision models, vector re-weighting is not that necessary, which implies the nonlinearity is primarily for scalar features. In that case, vector mapping preserves structural information while scalar updating provides feature discrimination. In contrast, for binary structures, both interactions are needed.



\vspace{0.3em}
\noindent \textbf{Visualization} \quad
In Fig.~\ref{fig:visualization}, we gave qualitative evaluations by visualizing the part segmentation predictions based on PointNet. The original PointNet is rotation sensitive, which produces unstable predictions if the testing input is randomly rotated. As shown in the figure, the issue was desirably solved when configuring it via SVNet.

\vspace{0.3em}
\noindent \textbf{Future work} \quad
As we mentioned in the paper, techniques for building strong binary networks like knowledge distillation~\cite{hinton2015distilling}, learnable thresholds~\cite{zhang2022dynamicthreshold}, attentions~\cite{martinez2020realtobinary}, and multi-step training~\cite{martinez2020realtobinary} can further boost SVNet. Another potential future direction is to extend SVNet for transformers~\cite{vaswani2017attention,dosovitskiy2020visiontransformer,fuchs2020se3transformer} and graph convolutional neural networks (GCNs)~\cite{kipf2016semi} on regular/irregular data, which have been proved to be powerful, yet energy-consuming architectures. SVNet has a similar form to transformers or GCNs, \emph{i.e.}, information aggregation from orderless points (tokens in transformers) and point-wise updating (token-wise feed-forward updating in transformers). A rotation-robust property with high running efficiency is fundamentally desirable for such models for wider applications.

\section{Conclusion}

In this paper, we proposed a general 3D learning structure named SVNet that integrates binarization and SO(3) equivariance for point clouds. Binarization aims to enhance the network efficiency via constraining features and weights to \{-1, +1\}, so that most of the computation can be implemented with efficient binary operations rather than full-precision multiplications. Meanwhile, SO(3) equivariance enables the network to be robust to 3D rotations, which is the key for 3D point cloud models to adapt to environments with unseen poses. We analyzed the necessity of using both scalar and vector features, which leads us to obtain a better trade-off between efficiency, equivariance, and accuracy. 

\vspace{0.5em}
\noindent \textbf{Acknowledgement.} \quad 
This work was partially supported by National Key Research and Development Program of China No. 2021YFB3100800, the Academy of Finland under grant 331883 
and the National Natural Science Foundation of China under Grant 61872379 and 62022091. The CSC IT Center for Science, Finland, is also acknowledged for computational resources.

\appendix
\addcontentsline{toc}{section}{Appendices}
\section*{Appendices}

\section{Binarization algorithm in the experiments}

A binary linear function with weights $W$ mapping input $X$ to output $Y$ can be written as:
\begin{align}
    Y &= \text{Sign}(X)\cdot\text{Sign}(W)\; &\text{if both } X\text{ and }W\text{ are binarized},\label{eq:7}\\
    Y &= X\cdot\text{Sign}(W)\;&\text{if only }W \text{ is binarized},\label{eq:8}
\end{align}
where $\text{Sign}(a) = +1$ if $a \geq 0$ otherwise -1, ``$\cdot$'' means matrix multiplication. We use the shifting and scaling strategies~\cite{qin2020bipointnet,liu2020reactnet} to reduce the information loss:
\begin{equation}
    \underline{X = X-\beta} \to \underline{(\ref{eq:7})\; or\; (\ref{eq:8})}\to \underline{Y = \gamma Y},
\end{equation}
where $\beta$ and $\gamma$ are both channel-wise parameters and learned during training. We use the simplest Straight Through Estimator (STE)~\cite{hinton2012ste} with clipping range $(-1.2, 1.2)$\footnote{It should be noted we adopt a wider clipping range than $(-1, 1)$ for a better training performance.} to calculate the gradients for the Sign function such that the network can be trained with the standard gradient descent algorithm:

\begin{align}
    x^b &= \text{Sign}(x);\\
    g_x &= 
    \begin{cases}
    g_{x^b}&\text{if }-1.2<x<1.2,\\
    0\;&\text{otherwise}.
    \end{cases}
\end{align}
In SVNet, Eq.~\ref{eq:7} corresponds to the linear transformation of scalar features while Eq.~\ref{eq:8} the vector mapping of vector features.

\begin{table}[t!]
\begin{center}
\setlength{\tabcolsep}{0.005\linewidth}
\resizebox*{\linewidth}{!}{
\begin{tabular}{l|l|c|ccc|ccc}
\toprule
& Method & Params & MACs & ADDs & BOPs & I/$z$ & I/SO(3) & $z$/SO(3)\\
\hline
\multirow{4}{*}{DGCNN} & Original~\cite{wang2019dgcnn} & 57.7M & 2.4B & 0 & 0 & 37.2 & 16.6 & 33.8\\
& Vector Neurons~\cite{deng2021vn} & 92.8M & 3.2B & 0 & 0 & 90.0 & 90.0 & 89.5\\
& SVNet (FP) & 49.7M & 1.4B & 0 & 0 & 90.3 & 90.3 & 90.3\\
& \textbf{SVNet} & \textbf{3.4M} & \textbf{0.05B} & \textbf{0.2B} & \textbf{1.2B} & \textbf{83.5} & \textbf{83.5} & \textbf{83.8}\\
\midrule
\multirow{6}{*}{PointNet} & Original~\cite{qi2017pointnet}$^*$ & 111.1M & 0.4B & 0 & 0 & 25.4 & 9.2 & 17.0\\
& Vector Neurons$^*$~\cite{deng2021vn} & 63.1M & 2.0B & 0 & 0 & 84.9 & 84.9 & 85.6\\
& SVNet (FP) & 78.8M & 1.5B & 0 & 0 & 86.6 & 86.6 & 86.3\\
\hhline{~--------}
& BiPointNet$^*$~\cite{qin2020bipointnet} & 4.2M & 0.01B & 0 & 0.4B & 22.4 & 11.1 & 13.7\\ 
& \textbf{SVNet} & \textbf{8.7M} & \textbf{0.03B} & \textbf{0.2B} & \textbf{1.2B} & \textbf{76.1} & \textbf{76.1} & \textbf{76.3}\\
& \textbf{SVNet-small} & \textbf{4.3M} & \textbf{0.02B} & \textbf{0.03B} & \textbf{0.2B} & \textbf{66.4} & \textbf{66.4} & \textbf{66.4}\\
\bottomrule
\end{tabular}
}
\end{center}
\caption{Complexity comparison on ModelNet40. The numbers show overall accuracies (\%), Params (memory storage of the model) were recorded in bits. We also constructed SVNet-small to compare with BiPointNet with similar complexity. ``$*$'' indicates our implementations using the code provided by the authors}
\label{tab:modelnet40bops}
\end{table}

\begin{table}[t!]
\begin{center}
\setlength{\tabcolsep}{0.005\linewidth}
\resizebox*{\linewidth}{!}{
\begin{tabular}{l|l|c|ccc|ccc}
\toprule
& Method & Params & MACs & ADDs & BOPs & I/$z$ & I/SO(3) & $z$/SO(3)\\
\hline
\multirow{4}{*}{DGCNN} & Original~\cite{wang2019dgcnn} & 46.7M & 4.4B & 0 & 0 & 43.8 & 36.1 & 49.3\\
& Vector Neurons~\cite{deng2021vn} & 41.8M & 6.6B & 0 & 0 & 81.5 & 81.5 & 81.4\\
& SVNet (FP) & 43.2M & 7.2B & 0 & 0 & 81.4 & 81.4 & 81.4\\
& \textbf{SVNet} & \textbf{4.0M} & \textbf{0.2B} & \textbf{1.0B} & \textbf{6.0B} & \textbf{68.4} & \textbf{68.4} & \textbf{68.4}\\
\midrule
\multirow{6}{*}{PointNet} & Original$^*$~\cite{qi2017pointnet} & 267.0M & 5.8B & 0 & 0 & 38.5 & 33.0 & 41.8\\
& Vector Neurons$^*$~\cite{deng2021vn} & 162.6M & 20.5B & 0 & 0 & 79.8 & 79.8 & 79.8\\
& SVNet (FP) & 234.8M & 14.2B & 0 & 0 & 78.2 & 78.2 & 78.2 \\
\hhline{~--------}
& BiPointNet$^*$~\cite{qin2020bipointnet} & 9.0M & 0.1B & 0 & 5.7B & 34.5 & 29.7 & 31.4\\ 
& \textbf{SVNet} & \textbf{14.0M} & \textbf{0.2B} & \textbf{0.2B} & \textbf{13.8B} & \textbf{67.7} & \textbf{67.7} & \textbf{67.3}\\
& \textbf{SVNet-small} & \textbf{7.2M} & \textbf{0.1B} & \textbf{0.2B} & \textbf{5.2B} & \textbf{63.9} & \textbf{63.9} & \textbf{64.3}\\
\bottomrule
\end{tabular}
}
\end{center}
\caption{Complexity comparison on ShapeNet with corresponding mIoU (\%) over all classes. Other notes are the same as in Table~\ref{tab:modelnet40bops}}
\label{tab:shapenetbops}
\end{table}

\section{Rigorous rotation equivariance}

In order to demonstrate that our method achieves rigorous rotation equivariance, we evaluated it with the more challenging training/test settings I/$z$ and I/SO3, where the model was trained without any rotation augmentation, but tested with arbitrary rotations. The results in Table~\ref{tab:modelnet40bops} and Table~\ref{tab:shapenetbops} show that the outputs of SVNets are independent on rotations.

{\small
\bibliographystyle{ieee_fullname}
\bibliography{egbib}
}

\end{document}